\documentclass[10pt]{article}
\usepackage[margin=1.8in]{geometry}
\usepackage[hide links]{hyperref}

\usepackage{amsmath}
\usepackage{booktabs}
\usepackage{graphicx}
\usepackage{authblk}
\usepackage{palatino}

\newcommand{\etal}[0]{\textit{et al.}}
\newcommand{\jemph}[1]{\textsc{#1}}

\usepackage[font=small]{caption}
\usepackage{enumitem}
\date{}

\title{{\textbf{The Social Dynamics of Language Change in Online Networks}}\footnote{This paper appears in the Proceedings of the International Conference on Social Informatics (SocInfo16). The final publication is available at \url{springer.com}.}}
\author[1]{Rahul Goel}
\author[1]{Sandeep Soni}
\author[1]{Naman Goyal}
\author[2]{John Paparrizos}
\author[3]{Hanna Wallach}
\author[3]{Fernando Diaz}
\author[1]{Jacob Eisenstein}
\affil[1]{Georgia Institute of Technology, Atlanta, GA}
\affil[2]{Columbia University, New York, NY}
\affil[3]{Microsoft Research, New York, NY}

\newcommand{\ex}[1]{\textit{#1}}

\begin{document}
\maketitle

\begin{abstract}
Language change is a complex social phenomenon, revealing pathways of
communication and sociocultural influence.  But, while language change
has long been a topic of study in sociolinguistics, traditional
linguistic research methods rely on circumstantial evidence,
estimating the direction of change from differences between older and
younger speakers.  In this paper, we use a data set of several million
Twitter users to track language changes in progress.  First, we show
that language change can be viewed as a form of social influence: we
observe complex contagion for phonetic spellings and ``netspeak''
abbreviations (e.g., \emph{lol}), but not for older dialect markers
from spoken language. Next, we test whether specific types of social
network connections are more influential than others, using a
parametric Hawkes process model. We find that tie strength plays an
important role: densely embedded social ties are significantly better
conduits of linguistic influence. Geographic locality appears to play
a more limited role: we find relatively little evidence to support the
hypothesis that individuals are more influenced by geographically
local social ties, even in their usage of geographical dialect markers.
\end{abstract}

\section{Introduction}

Change is a universal property of language. For example, English has
changed so much that Renaissance-era texts like \emph{The Canterbury
  Tales} must now be read in translation. Even contemporary American
English continues to change and diversify at a rapid pace---to such an
extent that some geographical dialect differences pose serious
challenges for
comprehensibility~\cite{labov2011principles}. Understanding language
change is therefore crucial to understanding language itself, and has
implications for the design of more robust natural language processing
systems~\cite{eisenstein2013bad}.

Language change is a fundamentally social
phenomenon~\cite{labov2001principles}. For a new linguistic form to
succeed, at least two things must happen: first, speakers (and
writers) must come into contact with the new form; second, they must
decide to use it. The first condition implies that language change is
related to the structure of social networks. If a significant number
of speakers are isolated from a potential change, then they are
unlikely to adopt it~\cite{milroy1992social}. But mere exposure is not
sufficient---we are all exposed to language varieties that are
different from our own, yet we nonetheless do not adopt them in our
own speech and writing. For example, in the United States, many
African American speakers maintain a distinct dialect, despite being
immersed in a linguistic environment that differs in many important
respects~\cite{green2002african,rickford2010geographical}. Researchers
have made a similar argument for socioeconomic language differences in
Britain~\cite{trudgill1972sex}. In at least some cases, these
differences reflect questions of identity: because language is a key
constituent in the social construction of group identity, individuals
must make strategic choices when deciding whether to adopt new
linguistic
forms~\cite{bucholtz2005identity,johnstone2002dahntahn,labov1963social}. By
analyzing patterns of language change, we can learn more about the
latent structure of social organization: to whom people talk, and how
they see themselves.

But, while the basic outline of the interaction between language
change and social structure is understood, the fine details are still
missing: What types of social network connections are most important
for language change? To what extent do considerations of identity
affect linguistic differences, particularly in an online context?
Traditional sociolinguistic approaches lack the data and the methods
for asking such detailed questions about language variation and
change.

In this paper, we show that large-scale social media data can shed new
light on how language changes propagate through social networks. We
use a data set of Twitter users that contains all public messages for
several million accounts, augmented with social network and
geolocation metadata. This data set makes it possible to track, and
potentially explain, every usage of a linguistic variable\footnote{The
  basic unit of linguistic differentiation is referred to as a
  ``variable'' in the sociolinguistic and dialectological
  literature~\cite{wolfram1991linguistic}. We maintain this
  terminology here.} as it spreads through social media. 
Overall, we make the following contributions:

\begin{itemize}[leftmargin=0.2in]

\item We show that non-standard words are most likely to propagate
  between individuals who are connected in the Twitter mutual-reply
  network. This validates the basic approach of using Twitter to
  measure language change.

\item For some classes of non-standard words, we observe complex
  contagion---i.e., multiple exposures increase the likelihood of
  adoption. This is particularly true for phonetic spellings and
  ``netspeak'' abbreviations. In contrast, non-standard words that
  originate in speech do not display complex contagion.
  
\item We use a parametric Hawkes process
  model~\cite{hawkes1971spectra,li2014learning} to test whether
  specific types of social network connections are more influential
  than others. For some words, we find that densely embedded social
  ties are significantly better conduits of linguistic influence. This
  finding suggests that individuals make social evaluations of their
  exposures to new linguistic forms, and then use these social
  evaluations to strategically govern their own language use.
  
\item We present an efficient parameter estimation method that uses
  sparsity patterns in the data to scale to social networks with
  millions of users.

\end{itemize}

\section{Data}
\label{sec:data}

Twitter is an online social networking platform.  Users post
140-character messages, which appear in their followers' timelines.
Because follower ties can be asymmetric, Twitter serves multiple
purposes: celebrities share messages with millions of followers, while
lower-degree users treat Twitter as a more intimate social network for
mutual communication~\cite{kwak2010twitter}. In this paper, we use a
large-scale Twitter data set, acquired via an agreement between
Microsoft and Twitter. This data set contains all public messages
posted between June 2013 and June 2014 by several million users,
augmented with social network and geolocation metadata. We excluded
retweets, which are explicitly marked with metadata, and focused on
messages that were posted in English from within the United States.

\subsection{Linguistic Markers}
\label{sec:data-language}

The explosive rise in popularity of social media has led to an
increase in linguistic diversity and
creativity~\cite{androutsopoulos2011language,anis2007neography,baldwin2013noisy,crystal2006language,eisenstein2013bad,herring2012grammar},
affecting written language at all levels, from
spelling~\cite{eisenstein2015systematic} all the way up to grammatical
structure~\cite{tagliamonte2008linguistic} and semantic meaning across
the
lexicon~\cite{hamilton2016diachronic,kulkarni2015statistically}. Here,
we focus on the most easily observable and measurable level: variation
and change in the use of individual words.

We take as our starting point words that are especially characteristic
of eight cities in the United States. We chose these cities to
represent a wide range of geographical regions, population densities,
and demographics. We identified the following words as geographically
distinctive markers of their associated cities, using
SAGE~\cite{eisenstein2011sparse}. Specifically, we followed the
approach previously used by Eisenstein to identify community-specific
terms in textual
corpora~\cite{eisenstein2016geographical}.\footnote{After running SAGE
  to identify words with coefficients above $2.0$, we manually removed
  hashtags, named entities, non-English words, and descriptions of
  events.}
\begin{description}

\item[Atlanta:] \ex{ain} (phonetic spelling of \ex{ain't}), \ex{dese}
  (phonetic spelling of \ex{these}), \ex{yeen} (phonetic spelling of
  \ex{you ain't});

\item[Baltimore:] \ex{ard} (phonetic spelling of \ex{alright}),
  \ex{inna} (phonetic spelling of \ex{in a} and \ex{in the}), \ex{lls}
  (\ex{laughing like shit}), \ex{phony} (fake);

\item[Charlotte:] \ex{cookout};

\item[Chicago:] \ex{asl} (phonetic spelling of \ex{as hell}, typically
  used as an intensifier on Twitter\footnote{Other sources, such as
    \url{http://urbandictionary.com}, report \ex{asl} to be an
    abbreviation of \ex{age, sex, location?} However, this definition
    is not compatible with typical usage on Twitter, e.g.,
    \ex{currently hungry asl} or \ex{that movie was funny asl}.}),
  \ex{mfs} (\ex{motherfuckers});
  
\item[Los Angeles:] \ex{graffiti}, \ex{tfti} (\ex{thanks for the
  information});

\item[Philadelphia:] \ex{ard} (phonetic spelling of \ex{alright}),
  \ex{ctfuu} (expressive lengthening of \ex{ctfu}, an abbreviation of \ex{cracking the fuck up}),
  \ex{jawn} (generic noun);

\item[San Francisco:] \ex{hella} (an intensifier);

\item[Washington D.C.:] \ex{inna} (phonetic spelling of \ex{in a} and
  \ex{in the}), \ex{lls} (\ex{laughing like shit}), \ex{stamp} (an
  exclamation indicating emphasis).\footnote{\ex{ard}, \ex{inna}, and
    \ex{lls} appear on multiple cities' lists.  These words are
    characteristic of the neighboring cities of Baltimore,
    Philadelphia, and Washington D.C.}

\end{description}

Linguistically, we can divide these words into three main classes:
\begin{description}

\item[Lexical words:] The origins of \ex{cookout}, \ex{graffiti},
  \ex{hella}, \ex{phony}, and \ex{stamp} can almost certainly be
  traced back to spoken language. Some of these words (e.g.,
  \ex{cookout} and \ex{graffiti}) are known to all fluent English
  speakers, but are preferred in certain cities simply as a matter of
  topic. Other words (e.g., \ex{hella}~\cite{bucholtz2007hella} and
  \ex{jawn}~\cite{alim2009hip}) are dialect markers that are not
  widely used outside their regions of origin, even after several
  decades of use in spoken language.
  
\item[Phonetic spellings:] \ex{ain}, \ex{ard}, \ex{asl}, \ex{inna}, and
  \ex{yeen} are non-standard spellings that are based on phonetic variation by region, demographics, or situation.

\item[Abbreviations:] \ex{ctfuu}, \ex{lls}, \ex{mfs}, and \ex{tfti}
  are phrasal abbreviations.  These words are interesting because they
  are fundamentally textual.  They are unlikely to have come from
  spoken language, and are intrinsic to written social media.

\end{description}

Several of these words were undergoing widespread growth in popularity
around the time period spanned by our data set.  For example, the
frequencies of \ex{ard}, \ex{asl}, \ex{hella}, and \ex{tfti} more than
tripled between 2012 and 2013.  Our main research question is whether
and how these words spread through Twitter.  For example, lexical
words are mainly transmitted through speech. We would expect their
spread to be only weakly correlated with the Twitter social network.
In contrast, abbreviations are fundamentally textual in nature, so we
would expect their spread to correlate much more closely with the
Twitter social network.


\subsection{Social network}
\label{sec:data-social}

To focus on communication between peers, we constructed a social
network of mutual replies between Twitter users. Specifically, we
created a graph in which there is a node for each user in the data
set.  We then placed an undirected edge between a pair of users if
each replied to the other by beginning a message with their username.
Our decision to use the reply network (rather than the follower
network) was a pragmatic choice: the follower network is not widely
available. However, the reply network is also well supported by
previous research. For example, Huberman \etal\ argue that Twitter's
mention network is more socially meaningful than its follower network:
although users may follow thousands of accounts, they interact with a
much more limited set of users~\cite{huberman2008social}, bounded by a
constant known as Dunbar's number~\cite{dunbar1992neocortex}.
Finally, we restricted our focus to mutual replies because there are a
large number of unrequited replies directed at celebrities.  These
replies do not indicate a meaningful social connection.

\begin{figure}[t]
  \centering
  \includegraphics[width=0.5\textwidth]{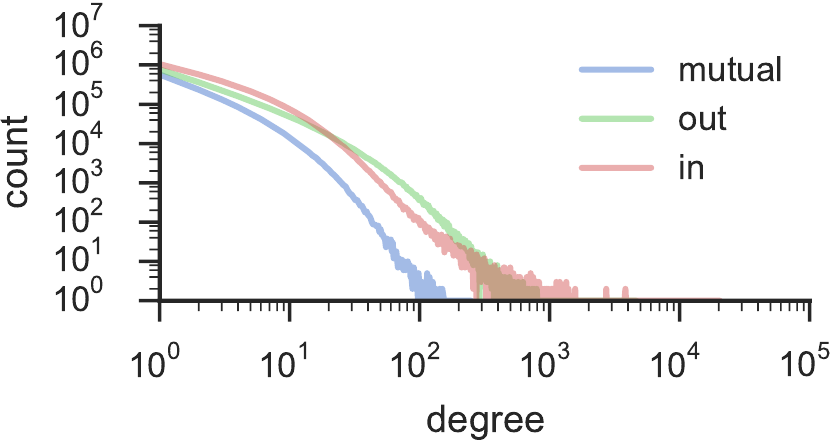}
  \caption{Degree distributions for our mutual-reply network and
    ``in'' and ``out'' networks.}
  \label{fig:degree-dist}
\end{figure}

We compared our mutual-reply network with two one-directional ``in''
and ``out'' networks, in which all public replies are represented by
directed edges.  The degree distributions of these networks are
depicted in \autoref{fig:degree-dist}.  As expected, there are a few
celebrities with very high in-degrees, and a maximum in-degree of
$20,345$.  In contrast, the maximum degree in our mutual-reply network
is $248$.

\subsection{Geography}
\label{sec:data-geography}

In order to test whether geographically local social ties are a
significant conduit of linguistic influence, we obtained geolocation
metadata from Twitter's location field.  This field is populated via a
combination of self reports and GPS tagging.  We aggregated metadata
across each user's messages, so that each user was geolocated to the
city from which they most commonly post messages. Overall, our data
set contains 4.35 million geolocated users, of which 589,562 were
geolocated to one of the eight cities listed in
\autoref{sec:data-language}. We also included the remaining users in
our data set, but were not able to account for their geographical
location.

Researchers have previously shown that social network connections in
online social media tend to be geographically
assortative~\cite{backstrom2010find,sadilek2012finding}.  Our data set
is consistent with this finding: for 94.8\% of mutual-reply dyads in
which both users were geolocated to one of the eight cities listed in
\autoref{sec:data-language}, they were both geolocated to the same
city. This assortativity motivates our decision to estimate separate
influence parameters for local and non-local social connections (see
\autoref{sec:parametric-hawkes}).

\section{Language Change as Social Influence}
\label{sec:influence}

Our main research goal is to test whether and how geographically
distinctive linguistic markers spread through Twitter.  With this goal
in mind, our first question is whether the adoption of these markers
can be viewed as a form of \jemph{complex contagion}.  To answer this
question, we computed the fraction of users who used one of the words
listed in \autoref{sec:data-language} after being exposed to that word
by one of their social network connections.  Formally, we say that
user $i$ \jemph{exposed} user $j$ to word $w$ at time $t$ if and only
if the following conditions hold: $i$ used $w$ at time $t$; $j$ had
not used $w$ before time $t$; the social network connection $i
\leftrightarrow j$ was formed before time $t$.  We define the
\jemph{infection risk} for word $w$ to be the number of users who use
word $w$ after being exposed divided by the total number of users who
were exposed.  To consider the possibility that multiple exposures
have a greater impact on the infection risk, we computed the infection
risk after exposures across one, two, and three or more distinct
social network connections.

The words' infection risks cannot be interpreted directly because
relational autocorrelation can also be explained by homophily and
external confounds.  For example, geographically distinctive
non-standard language is more likely to be used by young
people~\cite{pavalanathan2015confounds}, and online social network
connections are assortative by age~\cite{al2012homophily}.  Thus, a
high infection risk can also be explained by the confound of age.  
We therefore used the shuffle test proposed by Anagnostopoulos \etal~\cite{anagnostopoulos2008influence}, which compares the observed
infection risks to infection risks under the null hypothesis that
event timestamps are independent. The null hypothesis infection risks
are computed by randomly permuting the order of word usage events.  If
the observed infection risks are substantially higher than the
infection risks computed using the permuted data, then this is
compatible with social influence.\footnote{The
  shuffle test assumes that the likelihood of two users forming a
  social network connection does not change over time. Researchers
  have proposed a test~\cite{la2010randomization} that removes this
  assumption; we will scale this test to our data set in future work.}

\begin{figure}[t]
  \centering
  \includegraphics[width=0.8\textwidth]{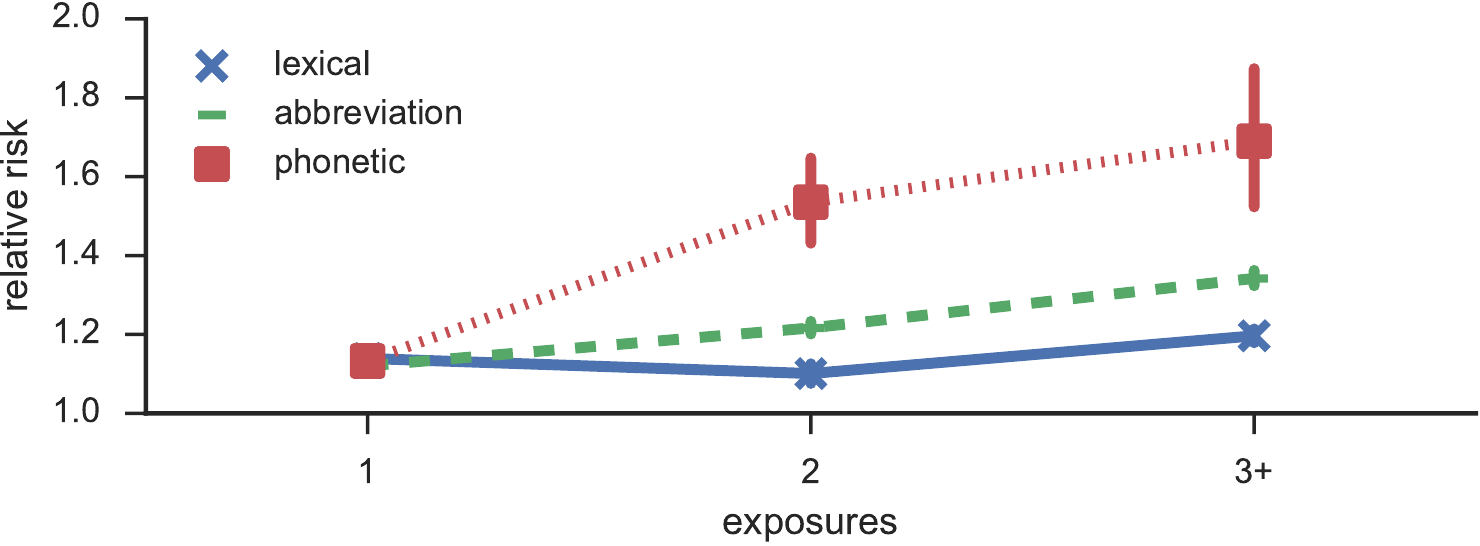}
  \caption{Relative infection risks for words in each of the three
    linguistic classes defined in \autoref{sec:data-language}. The
    figure depicts 95\% confidence intervals, computed using the
    shuffle test~\cite{anagnostopoulos2008influence}.}
  \label{fig:risk-by-exposure}
\end{figure}

\autoref{fig:risk-by-exposure} depicts the ratios between the words'
observed infection risks and the words' infection risks under the null
hypothesis, after exposures across one, two, and three or more
distinct connections.  We computed 95\% confidence intervals across
the words and across the permutations used in the shuffle test.  For
all three linguistic classes defined in \autoref{sec:data-language},
the risk ratio for even a single exposure is significantly greater
than one, suggesting the existence of social influence.  The risk
ratio for a single exposure is nearly identical across the three
classes. For phonetic spellings and abbreviations, the risk ratio
grows with the number of exposures.  This pattern suggests that words
in these classes exhibit \jemph{complex contagion}---i.e., multiple
exposures increase the likelihood of
adoption~\cite{centola2007complex}.  In contrast, the risk ratio for
lexical words remains the same as the number of exposures increases,
suggesting that these words spread by simple contagion.

Complex contagion has been linked to a range of behaviors, from participation in collective political action to adoption of avant
garde fashion~\cite{centola2007complex}.  A common theme among these
behaviors is that they are not cost-free, particularly if the behavior
is not legitimated by widespread adoption.  In the case of linguistic
markers intrinsic to social media, such as phonetic spellings and
abbreviations, adopters risk negative social evaluations of their
linguistic competency, as well as their cultural
authenticity~\cite{squires2010enregistering}.  In contrast, lexical
words are already well known from spoken language and are thus less
socially risky.  This difference may explain why we do not observe complex contagion for lexical words.

\section{Social Evaluation of Language Variation}
\label{sec:model}

In the previous section, we showed that geographically distinctive
linguistic markers spread through Twitter, with evidence of complex
contagion for phonetic spellings and abbreviations. But, does each
social network connection contribute equally? Our second question is
therefore whether (1) strong ties and (2) geographically local ties
exert greater linguistic influence than other ties. If so, users must
socially evaluate the information they receive from these connections,
and judge it to be meaningful to their linguistic self-presentation. In this section, we outline two hypotheses regarding their relationships to linguistic influence.

\subsection{Tie Strength}

Social networks are often characterized in terms of strong and weak
ties~\cite{granovetter1973strength,milroy1992social}, with strong ties
representing more important social relationships. Strong ties are
often densely embedded, meaning that the nodes in question share many
mutual friends; in contrast, weak ties often bridge disconnected
communities. Bakshy \etal\ investigated the role of weak ties in
information diffusion, through resharing of URLs on
Facebook~\cite{bakshy2012role}. They found that URLs shared across
strong ties are more likely to be reshared. However, they also found
that weak ties play an important role, because users tend to have more
weak ties than strong ties, and because weak ties are more likely to be a source of new information. 
In some respects, language change is similar to traditional information diffusion scenarios, such as resharing of URLs. But, in contrast, language connects with personal identity on a much deeper level than a typical URL. 
As a result, strong, deeply embedded ties may play a greater role in enforcing community norms.

We quantify tie strength in terms of \jemph{embeddedness}.
Specifically, we use the normalized mutual friends metric introduced
by Adamic and Adar~\cite{adamic2003friends}:
\begin{equation}
s_{i,j} = \sum_{k \in \Gamma(i) \cap \Gamma(j)} \frac{1}{\log\left(
  \#| \Gamma(k)|\right)},
\end{equation}
where, in our setting, $\Gamma(i)$ is the set of users connected to
$i$ in the Twitter mutual-reply network and $\#|\Gamma(i)|$ is the
size of this set. This metric rewards dyads for having many mutual
friends, but counts mutual friends more if their degrees are low---a
high-degree mutual friend is less informative than one with a
lower-degree. Given this definition, we can form the following
hypothesis:
\begin{description}
\item[H1] The linguistic influence exerted across ties with a high
  embeddedness value $s_{i,j}$ will be greater than the linguistic
  influence exerted across other ties.
\end{description}

\subsection{Geographic Locality}

An open question in sociolinguistics is whether and how local
\jemph{covert prestige}---i.e., the positive social evaluation of
non-standard dialects---affects the adoption of new linguistic
forms~\cite{trudgill1972sex}. Speakers often explain their linguistic
choices in terms of their relationship with their local
identity~\cite{eckert2000language}, but this may be a post-hoc
rationalization made by people whose language is affected by factors
beyond their control.  Indeed, some sociolinguists have cast doubt on
the role of ``local games'' in affecting the direction of language
change~\cite{labov2002penelope}.

The theory of covert prestige suggests that geographically local social ties are more influential than non-local ties.  
We do not know of any prior attempts to test this hypothesis quantitatively. 
Although researchers have shown that local linguistic forms are more likely to be used in messages that address geographically local friends~\cite{pavalanathan2015linguistic}, they have not attempted to measure the impact of exposure to these forms.
This lack of prior work may be because it is difficult to obtain relevant data, and to make reliable inferences from such data.
For example, there are several possible explanations for the observation that people often use similar language to that of their geographical neighbors.
One is exposure: even online social ties tend to be geographically assortative~\cite{al2012homophily}, so most people are likely to be exposed to local linguistic forms through local ties.
Alternatively, the causal relation may run in the reverse direction, with individuals preferring to form social ties with people whose language matches their own.
In the next section, we describe a model that enables us to
tease apart the roles of geographic assortativity and local influence,
allowing us to test the following hypothesis:
\begin{description}
\item[H2] The influence toward geographically distinctive linguistic
  markers is greater when exerted across geographically local ties
  than across other ties.
\end{description}
We note that this hypothesis is restricted in scope to geographically
distinctive words. We do not consider the more general hypothesis that
geographically local ties are more influential for all types of
language change, such as change involving linguistic variables that
are associated with gender or socioeconomic status.

\section{Language Change as a Self-exciting Point Process}
\label{sec:point-process}

To test our hypotheses about social evaluation, we require a more
sophisticated modeling tool than the simple counting method described
in \autoref{sec:influence}. In this section, rather than asking
whether a user was previously exposed to a word, we ask by whom, in
order to compare the impact of exposures across different types of
social network connections. We also consider temporal properties. For
example, if a user adopts a new word, should we credit this to an
exposure from a weak tie in the past hour, or to an exposure from a
strong tie in the past day?

Following a probabilistic modeling approach, we treated our Twitter
data set as a set of cascades of timestamped events, with one cascade
for each of the geographically distinctive words described in
\autoref{sec:data-language}. Each event in a word's cascade
corresponds to a tweet containing that word. We modeled each cascade
as a probabilistic process, and estimated the parameters of this
process.  By comparing nested models that make progressively finer
distinctions between social network connections, we were able to
quantitatively test our hypotheses.

Our modeling framework is based on a \jemph{Hawkes process}~\cite{hawkes1971spectra}---a specialization of an inhomogeneous Poisson process---which explains a cascade of timestamped events in terms of influence parameters.
In a temporal setting, an inhomogeneous Poisson process says that the
number of events $y_{t_1,t_2}$ between $t_1$ and $t_2$ is drawn from a
Poisson distribution, whose parameter is the area under a time-varying
\jemph{intensity function} over the interval defined by $t_1$ and
$t_2$:
\begin{align}
y_{t_1,t_2} &\sim \text{Poisson}\left(\Lambda(t_1,t_2)\right))
\intertext{where}
\Lambda(t_1,t_2) &= \int_{t_1}^{t_2} \lambda(t)\ \textrm{d}t.
\end{align}
Since the parameter of a Poisson distribution must be non-negative,
the intensity function must be constrained to be non-negative for all
possible values of $t$.

A Hawkes process is a self-exciting inhomogeneous Poisson process, where the intensity function depends on previous events. 
If we have a cascade of $N$ events $\{t_n\}_{n=1}^N$, where $t_n$ is the timestamp of event $n$, then the intensity function is
\begin{equation}
\lambda(t) = \mu_t + \sum_{t_n < t} \alpha\, \kappa(t - t_n),
\label{eq:intensity_self}
\end{equation}
where $\mu_t$ is the base intensity at time $t$, $\alpha$ is an
influence parameter that captures the influence of previous events,
and $\kappa(\cdot)$ is a time-decay kernel.

We can extend this framework to vector observations
$\boldsymbol{y}_{t_1,t_2} = (y^{(1)}_{t_1, t_2}, \ldots, y^{(M)}_{t_1,
  t_2})$ and intensity functions $\boldsymbol{\lambda}(t) =
(\lambda^{(1)}(t), \ldots, \lambda^{(M)}(t))$, where, in our setting,
$M$ is the total number of users in our data set.  If we have a
cascade of $N$ events $\{(t_n, m_n)\}_{n=1}^N$, where $t_n$ is the
timestamp of event $n$ and $m_n \in \{1, \ldots, M\}$ is the source of
event $n$, then the intensity function for user $m' \in \{1, \ldots,
M\}$ is
\begin{equation}
\lambda^{(m')}(t) = \mu^{(m')}_t + \sum_{t_n < t} \alpha_{m_n \to m'} \kappa(t - t_n),
\label{eq:intensity}
\end{equation}
where $\mu_t^{(m')}$ is the base intensity for user $m'$ at time $t$,
$\alpha_{m_n \to m'}$ is a pairwise influence parameter that captures
the influence of user $m_n$ on user $m'$, and $\kappa(\cdot)$ is a
time-decay kernel. Throughout our experiments, we used an exponential
decay kernel $\kappa(\Delta t) = e^{-\gamma \Delta t}$. We set the
hyperparameter $\gamma$ so that $\kappa(\textrm{1 hour}) = e^{-1}$.

Researchers usually estimate all $M^2$ influence parameters of a
Hawkes process (e.g.,~\cite{li2014identifying,zhao2015seismic}).
However, in our setting, $M > 10^6$, so there are $O(10^{12})$
influence parameters.  Estimating this many parameters is
computationally and statistically intractable, given that our data set
includes only $O(10^5)$ events (see the $x$-axis of
\autoref{fig:ll-diffs} for event counts for each word).  Moreover,
directly estimating these parameters does not enable us to
quantitatively test our hypotheses.

\subsection{Parametric Hawkes Process}
\label{sec:parametric-hawkes}

Instead of directly estimating all $O(M^2)$ pairwise influence
parameters, we used Li and Zha's parametric Hawkes
process~\cite{li2014learning}. This model defines each pairwise
influence parameter in terms of a linear combination of pairwise
features:
\begin{equation}
\alpha_{m \to m'} = \boldsymbol{\theta}^{\top} \boldsymbol{f}(m \to m'),
\label{eq:alpha}
\end{equation}
where $\boldsymbol{f}(m \to m')$ is a vector of features that describe
the relationship between users $m$ and $m'$. Thus, we only need to
estimate the feature weights $\boldsymbol{\theta}$ and the base
intensities. To ensure that the intensity functions $\lambda^{(1)}(t),
\ldots, \lambda^{(M)}(t)$ are non-negative, we must assume that
$\boldsymbol{\theta}$ and the base intensities are non-negative.

We chose a set of four binary features that would enable us to test
our hypotheses about the roles of different types of social network
connections:
\begin{description}

\item[F1 Self-activation:] This feature fires when $m' \!=\! m$. We included this feature to capture the scenario where using a word once makes a user more likely to use it again, perhaps because they are adopting a non-standard style.

\item[F2 Mutual reply:] This feature fires if the dyad $(m, m')$ is in the Twitter mutual-reply network described in \autoref{sec:data-social}. 
We also used this feature to define the remaining two features. 
By doing this, we ensured that features F2, F3, and F4 were (at least) as sparse as the mutual-reply network.

\item[F3 Tie strength:] This feature fires if the dyad $(m,m')$ is in in the Twitter mutual-reply network, and the Adamic-Adar value for this dyad is especially high. 
Specifically, we require that the Adamic-Adar value be in the 90$^{\textrm{th}}$ percentile among all dyads where at least one user has used the word in question. Thus, this feature picks out the most densely embedded ties.
  
\item[F4 Local:] This feature fires if the dyad $(m,m')$ is in the Twitter mutual-reply network, and the users were geolocated to the same city, and that city is one of the eight cities listed in \autoref{sec:data}. For other dyads, this feature returns zero. Thus, this feature picks out a subset of the geographically local ties.

\end{description}

In \autoref{sec:results}, we describe how we used these features to
construct a set of nested models that enabled us to test our
hypotheses. In the remainder of this section, we provide the
mathematical details of our parameter estimation method.

\subsection{Objective Function}
\label{sec:model-likelihood}

We estimated the parameters using constrained maximum
likelihood. Given a cascade of events $\{(t_n, m_n)\}_{n=1}^N$, the
log likelihood under our model is
\begin{equation}
\mathcal{L} = \sum_{n=1}^N \log \lambda^{(m_n)}(t_n) - \sum_{m = 1}^M \int_0^T \lambda^{(m)}(t)\ \textrm{d}t,
\end{equation}
where $T$ is the temporal endpoint of the cascade. Substituting in the complete definition of the per-user intensity functions from~\autoref{eq:intensity} and \autoref{eq:alpha},
\begin{align}
\mathcal{L} &= \sum_{n=1}^N \log{\left(\mu^{(m_n)}_{t_n} + \sum_{t_{n'} < t_n} \boldsymbol{\theta}^{\top}\boldsymbol{f}(m_{n'} \to m_n)\,\kappa(t_n - t_{n'}) \right)} -{} \notag\\
&\quad \sum^M_{m'=1} \int_0^T \left(\mu_t^{(m')} + \sum_{t_{n'} < t} \boldsymbol{\theta}^{\top} \boldsymbol{f}(m_{n'} \to m')\, \kappa(t - {t_{n'}})\right)\textrm{d}t.
\end{align}
If the base intensities are constant with respect to time, then
\begin{align}
\mathcal{L} &= \sum_{n=1}^N \log{\left(\mu^{(m_n)} + \sum_{t_{n'} < t_n} \boldsymbol{\theta}^{\top}\boldsymbol{f}(m_{n'} \to m_n)\, \kappa(t_n - t_{n'}) \right)} - {}\notag\\
&\quad \sum^M_{m'=1} \left( T\mu^{(m')} + \sum^N_{n=1} \boldsymbol{\theta}^{\top} \boldsymbol{f}(m_n \to m')\,(1 - \kappa(T - t_n))\right),
\end{align}
where the second term includes a sum over all events $n = \{1, \ldots,
N\}$ that contibute to the final intensity $\lambda^{(m')}(T).$ To
ease computation, however, we can rearrange the second term around the
source $m$ rather than the recipient $m'$:
\begin{align}
\mathcal{L} &= \sum_{n=1}^N \log{\left(\mu^{(m_n)} + \sum_{t_{n'} < t_n} \boldsymbol{\theta}^{\top}\boldsymbol{f}(m_{n'} \to m_n)\, \kappa(t_n - t_{n'}) \right)} - \notag\\
&\quad \sum_{m=1}^M \left(T\mu^{(m)} + \sum_{\{n : m_n = m\}} \, \boldsymbol{\theta}^{\top} \boldsymbol{f}(m \to \star)\, (1 -  \kappa(T-t_n))\right),
\label{eq:ll-final}
\end{align}
where we have introduced an aggregate feature vector $\boldsymbol{f}(m
\to \star) = \sum_{m'=1}^M \boldsymbol{f}(m \to m')$.  Because the sum
$\sum_{\{n : m_n = m'\}} \boldsymbol{f}(m' \to \star)\,\kappa(T-t_n)$
does not involve either $\boldsymbol{\theta}$ or $\mu^{(1)}, \ldots,
\mu^{(M)}$, we can pre-compute it. Moreover, we need to do so only for
users $m \in \{1, \ldots, M\}$ for whom there is at least one event in
the cascade.

A Hawkes process defined in terms of \autoref{eq:intensity} has a log
likelihood that is convex in the pairwise influence parameters and
the base intensities. For a parametric Hawkes process, $\alpha_{m \to
  m'}$ is an affine function of $\boldsymbol{\theta}$, so, by
composition, the log likelihood is convex in $\boldsymbol{\theta}$ and
remains convex in the base intensities.

\subsection{Gradients}

The first term in the log likelihood and its gradient contains a
nested sum over events, which appears to be quadratic in the number of
events. However, we can use the exponential decay of the kernel
$\kappa(\cdot)$ to approximate this term by setting a threshold
$\tau^{\star}$ such that $\kappa(t_n - t_{n'}) = 0$ if $t_n - t_{n'}
\geq \tau^{\star}$. For example, if we set $\tau^{\star} = 24 \textrm{
  hours}$, then we approximate $\kappa(\tau^{\star}) = 3 \times
10^{-11} \approx 0$. This approximation makes the cost of computing
the first term linear in the number of events.

The second term is linear in the number of social network connections
and linear in the number of events. Again, we can use the exponential
decay of the kernel $\kappa(\cdot)$ to approximate $\kappa(T - t_n)
\approx 0$ for $T - t_n \geq \tau^{\star}$, where $\tau^{\star} = 24
\textrm{ hours}$. This approximation means that we only need to
consider a small number of tweets near temporal endpoint of the
cascade. For each user, we also pre-computed $\sum_{\{n : m_n = m'\}}
\boldsymbol{f}(m' \to \star)\,\kappa(T - t_n)$. Finally, both terms in
the log likelihood and its gradient can also be trivially parallelized over
users $m = \{1, \ldots, M\}$.

For a Hawkes process defined in terms of \autoref{eq:intensity}, Ogata
showed that additional speedups can be obtained by recursively
pre-computing a set of aggregate messages for each dyad $(m,
m')$. Each message represents the events from user $m$ that may
influence user $m'$ at the time $t_i^{(m')}$ of their
$i^{\textrm{th}}$ event~\cite{ogata1981lewis}:
\begin{align*}
&R^{(i)}_{m \to m'} \notag\\
  &\quad = 
\begin{cases}
\kappa(t^{(m')}_{i} - t^{(m')}_{i-1})\,R^{(i-1)}_{m \to m'} + \sum_{t^{(m')}_{i-1} \leq t^{(m)}_{j} \leq t^{(m')}_i} \kappa(t^{(m')}_i - t^{(m)}_j) & m\neq m'\\
\kappa(t^{(m')}_{i} - t^{(m')}_{i-1}) \times (1 + R^{(i-1)}_{m \to m'}) & m = m'.
\end{cases}
\end{align*}
These aggregate messages do not involve the feature weights
$\boldsymbol{\theta}$ or the base intensities, so they can be
pre-computed and reused throughout parameter estimation.

For a parametric Hawkes process, it is not necessary to compute a set
of aggregate messages for each dyad. It is sufficient to compute a set
of aggregate messages for each possible configuration of the
features. In our setting, there are only four binary features, and
some combinations of features are impossible.

Because the words described in \autoref{sec:data-language} are
relatively rare, most of the users in our data set never used them.
However, it is important to include these users in the model.  Because
they did not adopt these words, despite being exposed to them by users
who did, their presence exerts a negative gradient on the feature
weights.  Moreover, such users impose a minimal cost on parameter
estimation because they need to be considered only when pre-computing
feature counts.

\subsection{Coordinate Ascent}

We optimized the log likelihood with respect to the feature weights
$\boldsymbol{\theta}$ and the base intensities. Because the log
likelihood decomposes over users, each base intensity $\mu^{(m)}$ is
coupled with only the feature weights and not with the other base
intensities. Jointly estimating all parameters is inefficient because
it does not exploit this structure. We therefore used a coordinate
ascent procedure, alternating between updating $\boldsymbol{\theta}$
and the base intensities. As explained in
\autoref{sec:parametric-hawkes}, both $\boldsymbol{\theta}$ and the base
intensities must be non-negative to ensure that intensity functions
are also non-negative. At each stage of the coordinate ascent, we
performed constrained optimization using the active set method of
MATLAB's \texttt{fmincon} function.

\section{Results}
\label{sec:results}

We used a separate set of parametric Hawkes process models for each of
the geographically distinctive linguistic markers described in
\autoref{sec:data-language}.  Specifically, for each word, we
constructed a set of nested models by first creating a baseline model
using features F1 (self-activation) and F2 (mutual reply) and then
adding in each of the experimental features---i.e., F3 (tie strength)
and F4 (local).

We tested hypothesis H1 (strong ties are more influential) by
comparing the goodness of fit for feature set F1+F2+F3 to that of
feature set F1+F2.  Similarly, we tested H2 (geographically local ties
are more influential) by comparing the goodness of fit for feature set
F1+F2+F4 to that of feature set F1+F2.

\begin{figure}[t]
\centering
\includegraphics[width=\textwidth]{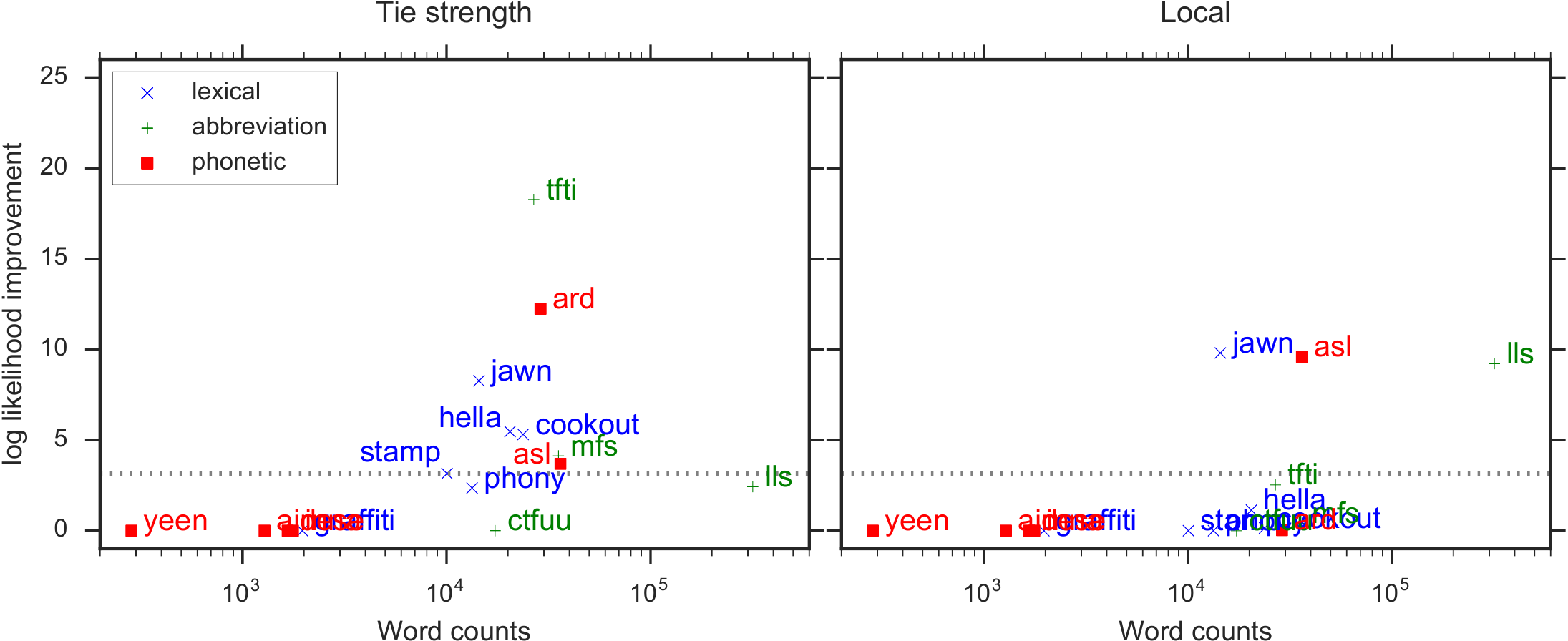}
\caption{Improvement in goodness of fit from adding in features F3
  (tie strength) and F4 (local). The dotted line corresponds to
  the threshold for statistical significance at $p<0.05$ using a
  likelihood ratio test with the Benjamini-Hochberg correction.}
\label{fig:ll-diffs}
\end{figure}

In \autoref{fig:ll-diffs}, we show the improvement in goodness of fit
from adding in features F3 and F4.\footnote{We also
  compared the full feature set---i.e., F1+F2+F3+F4---to feature set
  F1+F2+F3 and feature set F1+F2+F4. The results were almost
  identical, indicating that F3 (tie strength) and F4 (local) provide
  complementary information.}  
Under the null hypothesis, the log of the likelihood ratio follows a $\chi^2$ distribution with one degree of freedom, because the models differ by one parameter. 
Because we performed thirty-two hypothesis tests (sixteen words, two features), we needed to adjust the significance thresholds to correct for multiple comparisons. 
We did this using the Benjamini-Hochberg procedure~\cite{benjamini1995controlling}.

Features F3 and F4 did not improve the goodness of fit for less
frequent words, such as \ex{ain}, \ex{graffiti}, and \ex{yeen}, which
occur fewer than $10^4$ times. Below this count threshold, there is
not enough data to statistically distinguish between different types
of social network connections. However, above this count threshold,
adding in F3 (tie strength) yielded a statistically significant
increase in goodness of fit for \ex{ard}, \ex{asl}, \ex{cookout},
\ex{hella}, \ex{jawn}, \ex{mfs}, and \ex{tfti}. This finding provides
evidence in favor of hypothesis H1---that the linguistic influence
exerted across densely embedded ties is greater than the linguistic
influence exerted across other ties.

In contrast, adding in F4 (local) only improved goodness of fit for
three words: \ex{asl}, \ex{jawn}, and \ex{lls}.  
We therefore conclude that support for hypothesis H2---that the linguistic influence exerted across geographically local ties is greater than the linguistic influence across than across other ties---is limited at best.

In \autoref{sec:influence} we found that phonetic spellings and
abbreviations exhibit complex contagion, while lexical words do not.
Here, however, we found no such systematic differences between the
three linguistic classes.  Although we hypothesize that lexical words
propagate mainly outside of social media, we nonetheless see that when
these words do propagate across Twitter, their adoption is modulated
by tie strength, as is the case for phonetic spellings and abbreviations.

\section{Discussion}

Our results in \autoref{sec:influence} demonstrate that language
change in social media can be viewed as a form of information
diffusion across a social network.  Moreover, this diffusion is
modulated by a number of sociolinguistic factors.  For non-lexical
words, such as phonetic spellings and abbreviations, we find evidence
of complex contagion: the likelihood of their adoption increases with
the number of exposures.  For both lexical and non-lexical words, we
find evidence that the linguistic influence exerted across densely
embedded ties is greater than the linguistic influence exerted across
other ties.  In contrast, we find no evidence to support the
hypothesis that geographically local ties are more influential.

Overall, these findings indicate that language change is not merely a
process of random diffusion over an undifferentiated social network,
as proposed in many simulation
studies~\cite{fagyal2010centers,griffiths2007language,niyogi1997dynamical}.
Rather, some social network connections matter more than others, and
social judgments have a role to play in modulating language change.
In turn, this conclusion provides large-scale quantitative support for
earlier findings from ethnographic studies.  A logical next step would
be to use these insights to design more accurate simulation models,
which could be used to reveal long-term implications for language
variation and change.

Extending our study beyond North America is a task for future work.
Social networks vary dramatically across cultures, with traditional
societies tending toward networks with fewer but stronger
ties~\cite{milroy1992social}. The social properties of language
variation in these societies may differ as well.  Another important
direction for future work is to determine the impact of exogenous
events, such as the appearance of new linguistic forms in mass media.
Exogeneous events pose potential problems for estimating both
infection risks and social influence.  However, it may be possible to
account for these events by incorporating additional data sources,
such as search trends.  Finally, we plan to use our framework to study
the spread of terminology and ideas through networks of scientific
research articles.  
Here too, authors may make socially motivated decisions to adopt specific terms and ideas~\cite{latour2013laboratory}.  The principles behind these decisions might therefore be revealed by an analysis of linguistic events propagating over a social network.

\paragraph{Acknowledgments} Thanks to the reviewers for their feedback, to M\'arton Karsai for suggesting the infection risk analysis, and to Le Song for discussing Hawkes processes. John Paparrizos is an Alexander S. Onassis Foundation Scholar. This research was supported by the National Science Foundation under awards IIS-1111142 and RI-1452443, by the National Institutes of Health under award number R01-GM112697-01, and by the Air Force Office of Scientific Research.


\end{document}